\documentclass[a4paper,conference]{IEEEtran}
\IEEEoverridecommandlockouts

\usepackage{cite}
\usepackage{amsmath,amssymb,amsfonts}
\usepackage{graphicx}
\usepackage{textcomp}
\usepackage{xcolor}
\usepackage{tikz}
\usetikzlibrary{bayesnet}
\usetikzlibrary{arrows}
\usepackage{booktabs}
\usepackage[binary-units=true]{siunitx}
\usepackage{array,multirow}
\usepackage{hyperref}
\usepackage{enumitem}
\usepackage{algorithm}
\usepackage{graphicx}

\usepackage[noend]{algpseudocode}

\newcommand{\figref}[1]{Figure~\ref{#1}}

\newcolumntype{L}[1]{>{\raggedright\let\newline\\\arraybackslash\hspace{0pt}}m{#1}}
\newcolumntype{C}[1]{>{\centering\let\newline\\\arraybackslash\hspace{0pt}}m{#1}}
\newcolumntype{R}[1]{>{\raggedleft\let\newline\\\arraybackslash\hspace{0pt}}m{#1}}

\begin{document}

\title{VINNAS: Variational Inference-based Neural Network Architecture Search}

\author{\IEEEauthorblockN{Martin Ferianc\thanks{We thank Jakub Stano, George Punter and the reviewers for helpful feedback.}}
\IEEEauthorblockA{Dept. of Electronic and Electrical Engineering\\University College London\\\
WC1E 7JE, London, UK\\
Email: martin.ferianc.19@ucl.ac.uk}
\and
\IEEEauthorblockN{Hongxiang Fan}
\IEEEauthorblockA{Dept. of Computing\\
Imperial College London\\
SW7 2AZ, London, UK\\
Email: h.fan17@imperial.ac.uk}
\and
\IEEEauthorblockN{Miguel Rodrigues}
\IEEEauthorblockA{Dept. of Electronic and Electrical Engineering\\University College London\\\
WC1E 7JE, London, UK\\
Email: m.rodrigues@ucl.ac.uk}}

\maketitle

\begin{abstract}
    In recent years, neural architecture search (NAS) has received intensive scientific and industrial interest due to its capability of finding a neural architecture with high accuracy for various artificial intelligence tasks such as image classification or object detection. In particular, gradient-based NAS approaches have become one of the more popular approaches thanks to their computational efficiency during the search. However, these methods often experience a mode collapse, where the quality of the found architectures is poor due to the algorithm resorting to choosing a single operation type for the entire network, or stagnating at a local minima for various datasets or search spaces.
    To address these defects, we present a differentiable variational inference-based NAS method for searching sparse convolutional neural networks. Our approach finds the optimal neural architecture by dropping out candidate operations in an over-parameterised supergraph using variational dropout with automatic relevance determination prior, which makes the algorithm gradually remove unnecessary operations and connections without risking mode collapse. The evaluation is conducted through searching two types of convolutional cells that shape the neural network for classifying different image datasets. Our method finds diverse network cells, while showing state-of-the-art accuracy with up to almost $\boldsymbol{2 \times}$ fewer non-zero parameters.
\end{abstract}

\IEEEpeerreviewmaketitle

\section{Introduction}\label{sec:introduction}
Neural networks (NNs) have demonstrated their great potential in a wide range of artificial intelligence tasks such as image classification, object detection or speech recognition~\cite{zoph2016neural,  ding2020autospeech}. Nevertheless, designing a NN for a given task or a dataset requires significant human expertise, making their application restricted in the real-world~\cite{elsken2018neural}. Recently, neural architecture search (NAS) has been demonstrated to be a promising solution for this issue~\cite{zoph2016neural}, which automatically designs a NN for a given task on a target objective. Current NAS methods are already able to automatically find better neural architectures, in comparison to hand-made NNs~\cite{zoph2016neural, ding2020autospeech, real2019regularized}.

\begin{figure}[t]
    \centering
    \includegraphics[width=0.47\linewidth]{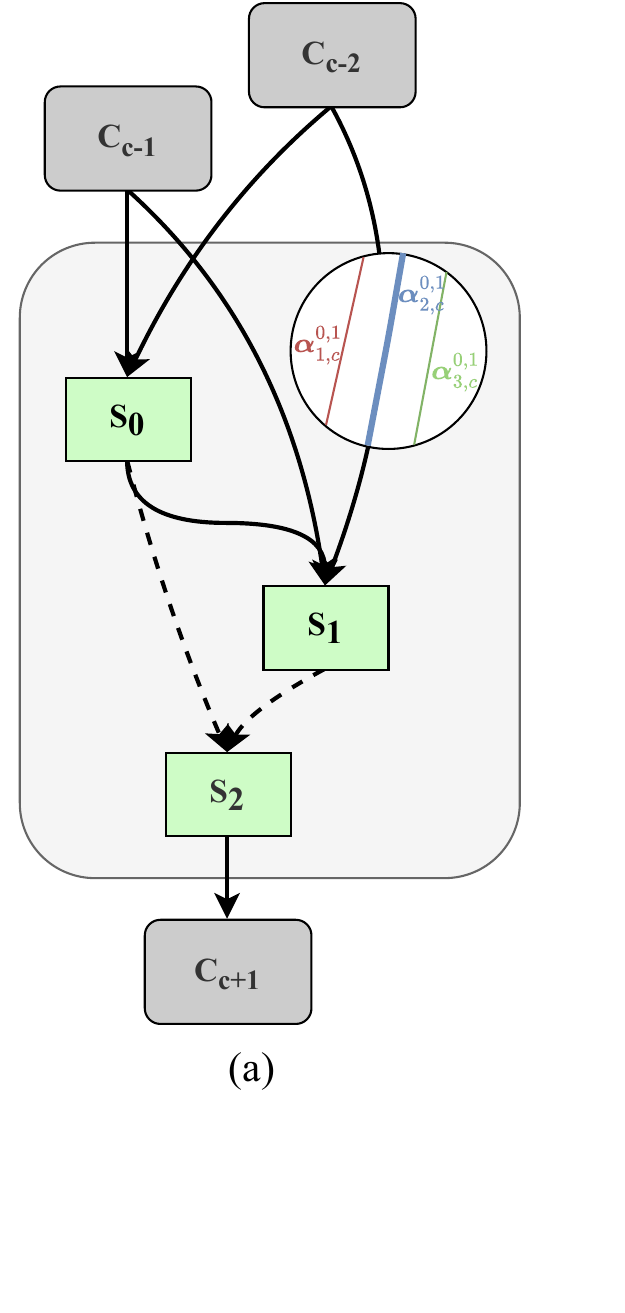}
    \hspace{1.5em}
    \includegraphics[height=0.37\textheight]{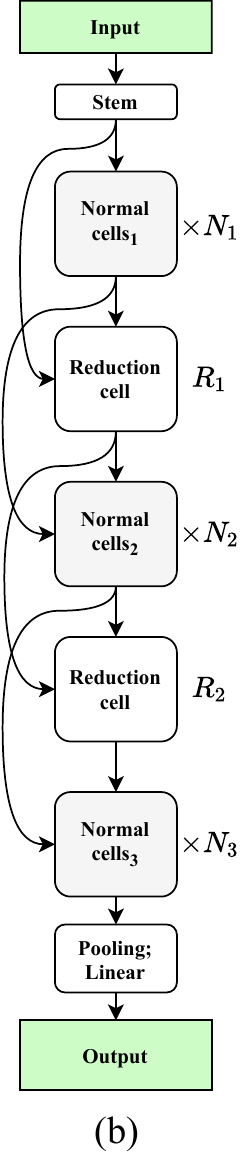}
\caption{(a) A structure of a computational cell which accepts inputs from two previous cells. Each cell accepts inputs from the immediate last cell $C_{c-1};c-1\geq0$ as well as the penultimate cell $C_{c-2};c-2\geq0$. The random variables $\boldsymbol{\alpha}$ represent the learnable relevance over the operations. The coloured lines represent different candidate operations and their thickness represents their likelihood. All outputs from the processed states $\boldsymbol{S}_0$ and $\boldsymbol{S}_1$ are concatenated in the output along the channel dimension into $C_{c+1}$, symbolised by the dashed lines. Green rectangles - states signify data. (b) Network skeleton comprising of $N_1, N_2$ and $N_3$ normal cells and two reduction cells $R_1$ and $R_2$ which share the same structure, in total giving $C=N_1 + N_2 + N_3 + 2$ cells. The network also contains a stem comprised of convolution and at the end of the network are average pooling followed by a linear classifier.}
\label{fig:nas_complete}
\end{figure}

NAS is a challenging optimisation problem on a constrained discrete search space, which can be simplified into reasoning about what operations should be present and how should they be interconnected between each other in the NN architecture. Common operation types considered in NAS are, for example, different types of convolutions or pooling~\cite{zoph2016neural}. However, if the search is not approached with caution, the resultant NN might not be flexible enough to learn useful patterns. Additionally, the ability of the model to generalise is also directly dependant on the NN architecture~\cite{zoph2016neural,liu2018darts}. Therefore, there is a pressing need for finding architectures that are expressive enough while achieving good generalisation performance.

Based on the core algorithmic principle operating during the search, NAS can be divided into four categories: (i) reinforcement learning-based on an actor-critic framework~\cite{zoph2016neural} (ii) evolutionary methods based on genetic algorithms~\cite{ real2019regularized}, (iii) Bayesian optimisation-based on proxy models~\cite{cai2018path} or (iv) gradient-based methods~\cite{liu2018darts}. In particular, gradient-based NAS~\cite{liu2018darts} has been recently popularised for convolutional NN (CNN) architecture search due to computational efficiency during the search. Nevertheless, gradient-based NAS is likely to collapse into a situation where it selects all operations to be the same~\cite{zela2019understanding}, treats operations unfairly~\cite{chu2019fair} or is hard to adapt across different datasets and search spaces~\cite{li2019random}.

To solve the issues in the existing gradient-based NAS methods, this paper proposes \textit{Variational Inference-based Neural Network Architecture Search (VINNAS)}. Under the same search space as in the case of NAS methods~\cite{liu2018darts,chu2020noisy,zela2019understanding}, our approach does not require any additional computation to the standard backpropagation algorithm. In \textit{VINNAS}, we tackle NAS using Bayesian inference, by modeling the architecture search through additional random variables $\boldsymbol{\alpha}$ which determine different operation types or connections between operations, our algorithm is able to conduct effective NN architecture search. The importance of using particular operations is determined by using a variational dropout scheme~\cite{molchanov2017variational, kingma2015variational} with the automatic relevance determination (ARD)~\cite{mackay1995probable} prior. We specifically search for a network structure that is composed of cells containing a variety of operations. The operations are organised into two types of cells: \textit{normal} and \textit{reduction}, and similarly to cell-based NAS~\cite{liu2018darts}, the cells are replicated and then used to construct the complete CNN. The model is shown in \figref{fig:nas_complete}. To encourage traversal through the NN architecture search space, we formulated an auto-regularising objective that promotes exploration, while ensuring high levels of certainty in the selection phase. 

We performed experiments on searching CNNs for classification on image datasets namely MNIST, FashionMNIST and CIFAR-10. Our results demonstrate the state-of-the-art (SOTA) performance, thanks to targeting
sparse architectures that focus on learning efficient representations, which is enforced by strict regularisation. For example on CIFAR-10, we demonstrate that our approach is able to find an architecture that contains $2 \times$ fewer non-zero parameters in comparison to the SOTA, without any human intervention.

In summary, our main contributions are as follows:
\begin{itemize}
    \item[1.] A differentiable neural architecture search method adopting variational dropout, which is effective in searching neural network architectures with the state-of-the-art performance on multiple datasets. 
    \item[2.] An architecture search objective using scheduled regularisation to promote exploration, but at the same time motivates certainty in the operation selection.
    \item[3.] An updated rule for selecting the most dominant operations based on their inferred uncertainty. 
\end{itemize}

In the sequel, we describe our approach in detail. In Section~\ref{sec:related_work} we review related work, in Section~\ref{sec:preliminaries} we introduce variational learning and gradient-based NAS. In Section \ref{sec:vinnas} we introduce our search objective, search space and the proposed overall algorithm. Section~\ref{sec:experiments} documents the performance of our search method on experiments and lastly, in Section~\ref{sec:conclusion} we draw our conclusions. 

\begin{table*}[t]
\centering
\caption{Notation used in this paper.}
\label{tb:notation}
\scalebox{.95}{
\setlength\tabcolsep{6pt} 
\begin{tabular}{ccccc}
\toprule
$\mathcal{A}$  Architecture & $\boldsymbol{\mathcal{M}}$  Architecture search space (supergraph) & $\boldsymbol{S}$  Data/State in architecture & $\boldsymbol{\alpha}$  Architecture var. & $\mathcal{D}/D$ Dataset / Dataset size  \\
$K$  Operation candidates & $o(.)$ Candidate operations & $C$ Total number of cells  &  N Normal cell & R Reduction cell \\ 
$p(.)$ Prior density & $q(.)$ Approximation density &  $\boldsymbol{w}$  Weights & $\boldsymbol{\Psi}$ Other params. &
$\boldsymbol{\theta}$  Reparametrisation params. \\
\bottomrule
\end{tabular}}
\end{table*}
We have since come across a competing publication~\cite{wang2020si} that overlaps with this work. In particular,~\cite{wang2020si} also proposes a NAS methodology for finding CNNs, based on ideas coming from variational dropout~\cite{kingma2015variational}. Additionally, the authors in~\cite{wang2020si} propose a hierarchical semi-implicit distribution over the operation as well as connectivity selection, that enables them to find CNN architectures with state-of-the-art accuracy. In our work, we imposed a distribution over the operation selection, while keeping the connectivity pattern fixed as shown in Figure~\ref{fig:nas_complete}, and the individual operation weights that allows us to find sparse and memory light-weight architectures.

\section{Related Work}\label{sec:related_work}
\paragraph{Differentiable Neural Architecture Search} Since Zoph \textit{et al.}~\cite{zoph2016neural} popularised NAS for CNNs, the field has been growing from intensive scientific~\cite{liu2018darts, zhou2019bayesnas} and industrial~\cite{zoph2016neural,real2019regularized} interests. NAS techniques automate the design of CNNs, mainly in terms of high-level operations, such as different types of convolutions or pooling, and their corresponding connections. The core of these techniques 
is the search space of potential architectures, their optimisation objective and search algorithm. For further detail of NAS, we refer the reader to a review of NAS by Elsken \textit{et al.}~\cite{elsken2018neural}. It is a common practice to organise the search space for all potential architectures into finding cells that specify the operations and their connections~\cite{liu2018darts}, which are then stacked on top of each other to construct the final NN, as previously shown in Figure~\ref{fig:nas_complete}. Modern NAS methods often apply a weight-sharing~\cite{pham2018efficient} approach where they optimise the search over several architectures in parallel by sharing weights of their operations to save memory consumption. 
Among these approaches, gradient-based NAS has become one of the most popular methods~\cite{liu2018darts}, mainly due to its compute feasibility.  DARTS~\cite{liu2018darts} defines the search for an architecture as optimising continuous weights associated to operations in an overparametrised supergraph $\boldsymbol{\mathcal{M}}$, while utilising weight-sharing. After the best combination of operations $\mathcal{A}; \mathcal{A} \subset \boldsymbol{\mathcal{M}}$ in the supergraph is identified, it is then used to construct the final architecture for evaluation. However, Zela \textit{et al.}~\cite{zela2019understanding} identified a wide range of search spaces for which DARTS yields degenerate architectures with very poor test performance. Chu \textit{et al.}~\cite{chu2019fair} observed critical problems in the two-stage weight-sharing NAS due to inherent unfairness in operation selection during the search in the supergraph. Chu \textit{et al.}~\cite{chu2020noisy} attempted to fix this problem by adding noise to the skip-connection operation during the search. Our approach is similar to~\cite{chu2020noisy}, however, we do not bias the search only towards skip-connections, but rather, infer the properties of the noise distribution with respect to ARD.

\paragraph{Pruning} Gradient-based NAS can be regarded as a subset of pruning in NNs, that is applied at the end of search in the operations' space. There have been many approaches introduced for pruning, such as by LeCun \textit{et al.}~\cite{lecun1990optimal} who pruned networks by analysing second-order derivatives. Other approaches~\cite{scardapane2017group} considered removing groups of filters in convolutions. Kingma \textit{et al.}~\cite{kingma2015variational} pruned NNs at a node-level by noticing connections between dropout~\cite{srivastava2014dropout} and variational inference. Molchanov \textit{et al.}~\cite{molchanov2017variational} showed that the interpretation of Gaussian dropout as performing variational inference in a network with log-uniform prior over weights leads to high sparsity in weights. Blundell \textit{et al.}~\cite{blundell2015weight} introduced a mixture of Gaussians prior on the weights, with one mixture tightly concentrated around zero, thus approximating a spike and slab prior over weights. Ghosh \textit{et al.}~\cite{ghosh2018structured} and Loizous \textit{et al.}~\cite{louizos2017bayesian} simultaneously considered grouped Horseshoe prior~\cite{carvalho2009handling} for neural pruning. Zhou \textit{et al.}~\cite{zhou2020posterior-guided} used variational dropout~\cite{kingma2015variational} to select filters for convolution. Our method differs to these approaches, by not only inferring sparse weights for operations, but also attempting to infer weights over the operations' search space to search NN architectures. 
\section{Preliminaries}\label{sec:preliminaries}
In this Section we introduce variational learning and cell-based differential neural architecture search which serve as basic building blocks for developing \textit{VINNAS}. Notation used in this paper is summarised in Table~\ref{tb:notation}.

\subsection{Variational Learning}\label{sec:preliminaries_variational_learning}
We specify a CNN as a parametrisable function approximator with some architecture $\mathcal{A}$ learnt on $D$ data samples consisting of inputs $\boldsymbol{x}_i$ and targets $\boldsymbol{y}_i$ forming a dataset $\mathcal{D}$ as $\mathcal{D}=\{(\boldsymbol{x_1}, \boldsymbol{y}_1), (\boldsymbol{x}_2, \boldsymbol{y}_2), (\boldsymbol{x}_3, \boldsymbol{y}_3), \ldots, (\boldsymbol{x}_D, \boldsymbol{y}_D)\}$. The architecture $\mathcal{A}$, composed of operations, might have certain parameters, for example weights $\boldsymbol{w}^{\mathcal{A}}$, which are distributed given some prior distributions $\boldsymbol{w}^{\mathcal{A}} \sim p(\boldsymbol{w})$. $\boldsymbol{w}^{\mathcal{A}}$ and $\mathcal{A}$ jointly define the model and the likelihood $p_{\mathcal{A}}(\boldsymbol{y}\mid\boldsymbol{x}, \boldsymbol{w}^{\mathcal{A}})$. We seek to learn the posterior distribution over the parameters $p_{\mathcal{A}}(\boldsymbol{w}^{\mathcal{A}} \mid \boldsymbol{x},\boldsymbol{y})$ using the Bayes rule. However, that is analytically intractable due to the normalising factor $p_{\mathcal{A}}(\boldsymbol{y} \mid \boldsymbol{x})$, which cannot be computed exactly due to the high dimensionality of $\boldsymbol{w}^{\mathcal{A}}$.

Therefore, we need to formulate an approximate parametrisable posterior distribution $q_{\mathcal{A}}(\boldsymbol{w}^{\mathcal{A}} \mid \boldsymbol{\theta}_w^\mathcal{A}, \boldsymbol{x},\boldsymbol{y})$\footnote{ From now on we drop the conditioning on the data $\{\boldsymbol{x},\boldsymbol{y}\}$ to avoid clutter in the notation, such that any parametrisable $q(.)$ will become $q(\boldsymbol{w} \mid \boldsymbol{\theta}_w)$.}
whose parameters $\boldsymbol{\theta}_w^\mathcal{A}$ can be learnt in order to approach the true posterior, $p_{\mathcal{A}}(\boldsymbol{w}^{\mathcal{A}} \mid \boldsymbol{x},\boldsymbol{y})$. Moving the distribution $q_{\mathcal{A}}(\boldsymbol{w}^{\mathcal{A}} \mid \boldsymbol{\theta}_w^\mathcal{A})$ closer to $p_{\mathcal{A}}(\boldsymbol{w}^{\mathcal{A}} \mid \boldsymbol{x},\boldsymbol{y})$ in terms of $\boldsymbol{\theta}_w^\mathcal{A}$ naturally raises an objective: to minimise their separation, which is expressed as the Kullback-Leibler ($\mathcal{KL}$) divergence~\cite{kullback1951information}. This objective $\mathcal{L}_\mathcal{A}(\boldsymbol{\theta}_w^\mathcal{A}, \boldsymbol{\Psi}^\mathcal{A})= \mathcal{KL}(q_{\mathcal{A}}(\boldsymbol{w}^{\mathcal{A}}\mid\boldsymbol{\theta}_w^\mathcal{A})\mid\mid p_{\mathcal{A}}(\boldsymbol{w}^{\mathcal{A}}\mid  \boldsymbol{x}, \boldsymbol{y}))$ is approximated through the evidence lower bound (ELBO), shown in~\eqref{eq:elbo}. The $\boldsymbol{\Psi}^\mathcal{A}$ represents other learnable pointwise parameters that are assumed to have uniform prior.
\begin{multline}
    \textrm{arg} \min_{\boldsymbol{\theta}_w^\mathcal{A},\boldsymbol{\Psi}^\mathcal{A}} \mathcal{KL}(q_{\mathcal{A}}(\boldsymbol{w}^{\mathcal{A}}\mid\boldsymbol{\theta}_w^\mathcal{A})\mid\mid p_{\mathcal{A}}(\boldsymbol{w}^{\mathcal{A}}\mid  \boldsymbol{x}, \boldsymbol{y})) = \\  = \textrm{arg} \min_{\boldsymbol{\theta}_w^\mathcal{A},\boldsymbol{\Psi}^\mathcal{A}} -\mathbb{E}_{q_{\mathcal{A}}(\boldsymbol{w}^{\mathcal{A}}\mid \ \boldsymbol{\theta}_w^\mathcal{A})}[\log p_{\mathcal{A}}(\boldsymbol{y} \mid \boldsymbol{x}, \boldsymbol{w}^{\mathcal{A}}, \boldsymbol{\Psi}^\mathcal{A})] + \\+\gamma \times \mathcal{KL}(q_{\mathcal{A}}(\boldsymbol{w}^{\mathcal{A}}\mid  \boldsymbol{\theta}_w^\mathcal{A}) \mid \mid p(\boldsymbol{w})) + const.
    \label{eq:elbo}
\end{multline}
The first term is the negative log-likelihood of the data which measures the data-fit, while the second term is a regulariser whose influence can be managed through $\gamma$. The $\boldsymbol{\Psi}^\mathcal{A}$ contribute to the $const.$ term that is independent of the parameters, due to the uniform prior. 

Kingma \textit{et al.} introduced the local reparametrisation trick (LRT)~\cite{kingma2015variational} that allows us to solve the objective in \eqref{eq:elbo} with respect to $\boldsymbol{\theta}_w^\mathcal{A}$ through stochastic gradient descent (SGD) with low variance. We can backpropagete the gradients with respect to the distribution $q_{\mathcal{A}}(\boldsymbol{w}^{\mathcal{A}}\mid \boldsymbol{\theta}_w^\mathcal{A})$ by sampling $\boldsymbol{z}$ that is obtained through deterministic transformation $t(.)$ as $\boldsymbol{z}=t(\boldsymbol{\theta}_w^\mathcal{A},\boldsymbol{\epsilon})$ where $\boldsymbol{\epsilon}$ is a parameter-free noise, e.g.: $\boldsymbol{\epsilon} \sim \mathcal{N}(\boldsymbol{0},\boldsymbol{I})$. 

Moreover, using this trick, Molchanov \textit{et al.}~\cite{molchanov2017variational}, were able to search for an unbounded approximation\footnote{$\odot$ represents a Hadamard product.} for weights $\boldsymbol{w}$ as shown in \eqref{eq:variational_dropout}, which corresponds to a Gaussian dropout model with learnable parameters $\boldsymbol{\theta}_w^\mathcal{A}=\{\boldsymbol{\mu}_w,\boldsymbol{\sigma}_w\}$~\cite{srivastava2014dropout}.
\begin{equation}
    \boldsymbol{w} \sim q_{\mathcal{A}}(\boldsymbol{w} \mid \boldsymbol{\mu}_w, \boldsymbol{\sigma}^2_w) \Leftrightarrow  \boldsymbol{w} = \boldsymbol{\mu}_w + \boldsymbol{\sigma}_w  \odot \boldsymbol{\epsilon}
    \label{eq:variational_dropout}
\end{equation}
After placing a factorised log-uniform prior on the weights, such that $p(\boldsymbol{w}) \propto \frac{1}{\mid \boldsymbol{w} \mid}$, the authors observed an effect similar to ARD~\cite{molchanov2017variational}, however, without the need to modify the prior. Throughout the inference, the learnt weights tend to a delta function centred at $\boldsymbol{0}$, leaving the model only with the important non-zero weights. The relevance determination is achieved by optimising both the $\boldsymbol{\mu}_w$ and $\boldsymbol{\sigma}_w$ and if they are both close to zero, they can be pruned. 

\subsection{Cell-based Differential Neural Architecture Search}\label{sec:preliminaries_nas_grad}
As shown above, Bayesian inference can be used to induce sparsity in the weight space, however, we wish to find $\mathcal{A}$ from some architecture space $\boldsymbol{\mathcal{M}};\mathcal{A} \subset \boldsymbol{\mathcal{M}}$.

Authors of DARTS~\cite{liu2018darts} defined the search for an architecture as finding specific $\boldsymbol{\alpha}$ associated to choosing operations $o(.)$ in an overparametrised directed acyclic graph (DAG) $\boldsymbol{\mathcal{M}}; \mathcal{A} \subset \boldsymbol{\mathcal{M}}$, where the learnt values of $\boldsymbol{\alpha}$ are then used to specify $\mathcal{A}$ at test time. Due to compute feasibility, the search space for all potential architectures is simplified into finding cells. The cell structure is defined with respect to $\boldsymbol{\alpha}; \boldsymbol{\alpha}^{i,j}_l \in \mathbb{R}^K; 1 \leq i < j<I$ where the indices $i,j$ signify the potential connections and operations $o_k(.)$ between information states $\boldsymbol{S}^{i}_c$ and $\boldsymbol{S}^{j}_c$ inside the cell $c$ with $I$ states, where $k\in 1,\ldots,K$. The information state $\boldsymbol{S}$ is a 4-dimensional tensor $\boldsymbol{S}\in \mathbb{R}^{B \times P \times H \times W}$ with $B$ samples, containing $P$ channels, height $H$ and width $W$. The index $l$ represents the number of different types of cells, where $l \in \{normal,reduce\}$ represents 2 different cell types: \textit{normal} (N) cells preserve the input dimensionality while \textit{reduce} (R) cells decrease the spatial dimensionality, but increase the number of channels~\cite{liu2018darts}. The cells can be interleaved and repeated giving $C$ total cells. The information for the state inside the cell $c$ is a weighted sum of the outputs generated from the $K$ different operations on $\boldsymbol{S}^{j}_c$. Choosing one of the operations can be approximated through performing $\text{softmax};\ \text{softmax}(\alpha^{i,j}_{l,k}) = \frac{\exp(\alpha^{i,j}_{l,k})}{\sum_{k^\prime} \exp(\alpha^{i,j}_{l,k^\prime})}$ on the architecture variables $\boldsymbol{\alpha}$, instead of argmax, which provides the method with differentiable strengths of potential operations as shown in \eqref{eq:darts}.
The last state $\boldsymbol{S}_{c}^{I}$, which is the output of the cell, is then a concatenation of all the previous states, except the first two input states $\boldsymbol{S}_{c}^I=\boldsymbol{S}_{c}^I\oplus\boldsymbol{S}_{c}^j; j<I$.
\begin{equation}
    \boldsymbol{S}^{i}_c = \sum_{j=1}^{j<i}\sum_{k=1}^{K} z_{c,k}^{i,j} o_{c,k}(\boldsymbol{S}^{j}_{c},\boldsymbol{w}^{i,j}_{c,k}) \quad \boldsymbol{z}_{c}^{i,j} = \textrm{softmax}(\boldsymbol{\alpha}_{l}^{i,j})
    \label{eq:darts}
\end{equation}
After the search, each state $\boldsymbol{S}_{c}^i$ is connected with the outputs from two operations $o_{c,k}^{j,l}(\boldsymbol{S}_c^j) + o_{c,k}^{i,l}(\boldsymbol{S}_c^i); i,j<l$, whose strengths $\boldsymbol{\alpha}$ have the highest magnitude. The learnt weights $\boldsymbol{w}$ are discarded and the resultant architecture is retrained from scratch. 

DARTS has been heavily adopted by the NAS community, due to its computational efficiency, in comparison to other NAS methods. However, upon a careful inspection, it can be observed that it does not promote choosing a particular operation and often collapses to a mode based on the fact that the graph is overparameterised through a variety of parallel operations~\cite{chu2019fair}. The supergraph then focuses on improving the performance with respect to the whole graph, without providing a dominant architecture. Additionally, others have observed~\cite{chu2019fair, chu2020noisy} that the method requires careful hyperparameter tuning without which it might collapse into preferring only one operation type over the others.

\section{VINNAS}\label{sec:vinnas}

In this Section, we first describe the search space assumptions for \textit{VINNAS} in detail, followed by the objective that guides the exploration among different architectures. At last, we present the algorithm of \textit{VINNAS} that couples everything together.

\subsection{Search Space}\label{sec:vinnas_search_space}
\begin{figure}[ht]
\centering
  \begin{tikzpicture}[scale=1.5]
  \node[obs] (y) {$y_i$};%
 \node[latent, left=of y] (x) {$\boldsymbol{x}_i$}; %
 \node[latent, right= of y] (w) {$\mathbf{w}$}; %
 \node[const, right= 0.5cm of w] (paramw) {$\boldsymbol{\mu}_w, \boldsymbol{\sigma}^2_w$};
 \node[latent, above=of y] (alphas) {$\boldsymbol{\alpha}$};
 \node[const, right= 0.5cm of alphas] (paramalphas_normal) {$\boldsymbol{\mu}_{normal}, \boldsymbol{\sigma}^2_{normal}$};
 \node[const, left= 0.5cm of alphas] (paramalphas_reduce) {$\boldsymbol{\mu}_{reduce}, \boldsymbol{\sigma}^2_{reduce}$};
 \plate[] {plate1} {(x)(y)} {$D$}; %

 \edge {paramalphas_normal, paramalphas_reduce}{alphas}
 \edge {x, w, alphas}{y}
 \edge{paramw}{w}
 \end{tikzpicture}
 
\caption{Graphical model capturing the search space in terms of the structural random variables $\boldsymbol{\alpha}$ and the weights $\boldsymbol{w}$. Note that, the parameters for $\boldsymbol{w}$ will be discarded after the search.}
 \label{fig:graphical_model}
\end{figure}
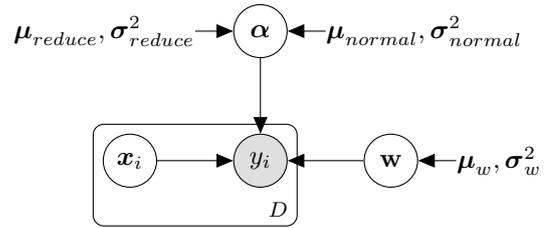

Our method extends the idea behind gradient-based NAS, while using variational learning to solve the aforementioned defects in previous work. \textit{VINNAS} builds its search space as an overparametrised DAG $\boldsymbol{\mathcal{M}}$ in which the algorithm searches for the right cell patterns to be used to build the final architecture $\mathcal{A}$. Similarly to DARTS, we aim to search for two repeated cells, namely a normal and a reduction cell that will be repeated as shown in \figref{fig:nas_complete}. Therefore, the $\boldsymbol{\mathcal{M}}$ contains several of normal and reduction cells laid in a sequence with each containing the $K$ parallel operation options. However, $\boldsymbol{\mathcal{M}}$ is downscaled in the number of cells and channels in comparison to the $\mathcal{A}$ considered during the evaluation, such that the supergraph can fit into GPU memory. Nevertheless, the pattern and the ratio of the number of cells $N_1, N_2$ and $N_3$ or $R$s in $\boldsymbol{\mathcal{M}}$ are preserved in accordance to the model shown in Figure~\ref{fig:nas_complete}. To apply variational inference and subsequently ARD through variational dropout, we associate the structural strength $\boldsymbol{\alpha}_{normal}$ for normal cells and $ \boldsymbol{\alpha}_{reduce}$ for reduction cells with a probabilistic interpretation. The graphical model of the supergraph $\boldsymbol{\mathcal{M}}$ that pairs together its weights $\boldsymbol{w}$ and architecture strengths $\boldsymbol{\alpha}$ is shown in \figref{fig:graphical_model}.

For simplicity, we assume fully factorisable log-uniform prior for $\boldsymbol{\alpha}= \{\boldsymbol{\alpha}_{normal}, \boldsymbol{\alpha}_{reduce}\}$. The prior biases the distributions of the operations' strengths towards zero, which avoids giving an advantage to certain operations over the others. We similarly model the weights $\boldsymbol{w}$ of the supergraph $\boldsymbol{\mathcal{M}}$ as random variables such that the joint prior distribution is $p(\boldsymbol{\alpha}, \boldsymbol{w})$ $=$ $ p(\boldsymbol{\alpha}_{normal}) p(\boldsymbol{\alpha}_{reduce})  p(\boldsymbol{w})$. It is not analytically possible to find the true posterior $p(\boldsymbol{\alpha}, \boldsymbol{w} \mid \boldsymbol{x}, \boldsymbol{y})$, therefore, we resort to formulating an approximation $q(\boldsymbol{\alpha}, \boldsymbol{w} \mid \boldsymbol{\theta}_\alpha, \boldsymbol{\theta}_w)$. We again set factorisable approximations for both $\boldsymbol{\alpha}$ and $\boldsymbol{w}$, such that the joint distribution factorises $q(\boldsymbol{\alpha}, \boldsymbol{w} \mid \boldsymbol{\theta}_\alpha, \boldsymbol{\theta}_w)$ $=$ $ q(\boldsymbol{\alpha}_{normal} \mid \boldsymbol{\theta}_{\alpha_{normal}}) q(\boldsymbol{\alpha}_{reduce} \mid \boldsymbol{\theta}_{\alpha_{reduce}})  q(\boldsymbol{w} \mid \boldsymbol{\theta}_w)$ with respect to the optimisable parameters $\boldsymbol{\theta}_w$ for $\boldsymbol{w}$ and $\boldsymbol{\theta}_\alpha=\{\boldsymbol{\theta}_{\alpha_{normal}},\boldsymbol{\theta}_{\alpha_{reduce}}\}$ for $\boldsymbol{\alpha}$. The prior $p(.)$ and approximations $q(.)$ are detailed in \eqref{eq:priors} and \eqref{eq:approxs} respectively. The indeces $i,j$ stand for different states in the cells with $i<j$ and $k$ is associated to the $K$ available operations.
\begin{align}
    p(\boldsymbol{w}) &=\prod_{i,j,k} p(\boldsymbol{w}_{k}^{i,j});\ p(\boldsymbol{w}_{k}^{i,j})\propto \frac{1}{\mid \boldsymbol{w}_{k}^{i,j} \mid} \label{eq:priors} \\
    p(\boldsymbol{\alpha}_{normal}) &=\prod_{i,j} p(\boldsymbol{\alpha}_{normal}^{i,j});\  p(\boldsymbol{\alpha}_{normal}^{i,j})  \propto \frac{1}{\mid \boldsymbol{\alpha}_{normal}^{i,j} \mid} \nonumber\\
    p(\boldsymbol{\alpha}_{reduce}) &=\prod_{i,j} p(\boldsymbol{\alpha}_{reduce}^{i,j});\  p(\boldsymbol{\alpha}_{reduce}^{i,j})  \propto \frac{1}{\mid \boldsymbol{\alpha}_{reduce}^{i,j}\mid} \nonumber
\end{align}
\begin{align}
     q(\boldsymbol{w}) &=\prod_{i,j,k} \mathcal{N}( \boldsymbol{\mu}^{i,j}_{w,k}, \boldsymbol{\sigma^2}_{w,k}^{i,j}) \label{eq:approxs} \\
     q(\boldsymbol{\alpha}_{normal}) &= \prod_{i,j} \mathcal{N}( \boldsymbol{\mu}^{i,j}_{\alpha_{normal}}, \boldsymbol{\sigma^2}^{i,j}_{\alpha_{normal}}) \nonumber \\
    q(\boldsymbol{\alpha}_{reduce}) &= \prod_{i,j} \mathcal{N}(\boldsymbol{\mu}^{i,j}_{\alpha_{reduce}},
    \boldsymbol{\sigma^2}^{i,j}_{\alpha_{reduce}}) \nonumber
\end{align}

The approximate posteriors were selected as Gaussians with diagonal covariance matrices. We used the formulation by Molchanov \textit{et al.}~\cite{molchanov2017variational} for both $\boldsymbol{\alpha}$, during the search phase, and $\boldsymbol{w}$, during both the search and test phases. We aim to induce sparsity in the operations' space, which would result in most operations' strengths in the DAG as zero, while the most relevant operations are expected to be non-zero. At the same time, the method induces sparsity in the weight space and thus motivates the individual operations to be extremely efficient in their learnt patterns.
Also, the Gaussian noise used in our method effectively disrupts the previously observed unfairness in operation selection during NAS as  partially demonstrated by~\cite{chu2020noisy} for skip-connection operation. Circling back to \eqref{eq:darts} the information in each cell during search is now calculated with respect to a sample $\boldsymbol{\alpha}$ from the inferred distributions $q(.)$. The second-level parameters such as the individual means and variances are assumed to have non-informative uniform prior.

\subsection{Search Objective}\label{sec:vinnas_search_objective}
The goal of the search is to determine the right set of structural variables $\boldsymbol{\alpha}$ or their corresponding parameters such that they can be later used to construct the desired architecture $\mathcal{A}$. Therefore, the search objective is to determine $\boldsymbol{\theta}_\alpha$ by solving $\mathcal{L}(\boldsymbol{\theta}_\alpha, \boldsymbol{\theta}_w, \boldsymbol{\Psi})$. $\mathcal{L}(\boldsymbol{\theta}_\alpha, \boldsymbol{\theta}_w, \boldsymbol{\Psi})$ is in fact a secondary objective to the primary objective of minimising~\eqref{eq:elbo} with respect to some unknown parameters implied by the chosen $\mathcal{A}$ as shown in \eqref{eq:min_min_formulation}.
\begin{equation}
    \textrm{arg} \min_{\boldsymbol{\theta}_{w}^\mathcal{A},\boldsymbol{\Psi}^{\mathcal{A}}, \boldsymbol{\theta}_\alpha} \mathcal{L}_{\mathcal{A}}( \boldsymbol{\theta}_{w}^\mathcal{A}, \boldsymbol{\Psi}^\mathcal{A}, \mathcal{L}(\boldsymbol{\theta}_\alpha, \boldsymbol{\theta}_w, \boldsymbol{\Psi}))
    \label{eq:min_min_formulation}
\end{equation}
The $\boldsymbol{\theta}_w, \boldsymbol{\theta}_\alpha$ and $\boldsymbol{\Psi}$ refer to the reparametrisations for the supergraph.
Therefore, at the same time it is necessary to optimise the objective with respect to the structural parameters $\boldsymbol{\theta}_\alpha$, the operations' weight parameters $\boldsymbol{\theta}_w$ and $\boldsymbol{\Psi}$ indicating their usefulness in the final architecture $\mathcal{A}$.  Derived from the original ELBO in \eqref{eq:elbo}, optimising the supergraph $\boldsymbol{\mathcal{M}}$ with respect to the learnable parameters rises the following objective in \eqref{eq:search_objective_1} below.
\begin{multline}
    \mathcal{A} \Leftarrow \boldsymbol{\theta}_{\boldsymbol{\alpha}}^{*} = \textrm{arg} \min_{\boldsymbol{\theta}_\alpha, \boldsymbol{\theta}_w, \boldsymbol{\Psi}} -\mathbb{E}_{q(\boldsymbol{\alpha},\boldsymbol{w})}[\log p(\boldsymbol{y}|\boldsymbol{x}, \boldsymbol{\alpha}, \boldsymbol{w}, \boldsymbol{\Psi})] + \\ + \gamma_1 \sum_{i,j,k,c} \mathcal{KL}(q(\boldsymbol{w}_{k,c}^{i,j}\mid \boldsymbol{\theta}_w) || p(\boldsymbol{w}_{k,c}^{i,j}))
    + \\ + \gamma_2 \sum_{i,j} \mathcal{KL}(q(\boldsymbol{\alpha}^{i,j}\mid \boldsymbol{\theta}_\alpha) || p(\boldsymbol{\alpha}^{i,j})) + const.
    \label{eq:search_objective_1}
\end{multline}

The first term again corresponds to the data-fitting term which pushes the parameters toward maximising the expectation of the log-likelihood with respect to the variational distributions $q(\boldsymbol{\alpha}, \boldsymbol{w} \mid \boldsymbol{\theta}_\alpha, \boldsymbol{\theta}_w)$ towards targets $\boldsymbol{y}$. The other two terms are regulariser terms, which because of the factorisation of the joint distributions $q(\boldsymbol{\alpha}, \boldsymbol{w})$ and priors $p(\boldsymbol{\alpha}, \boldsymbol{w})$ can be separated, and scaled by arbitrary constants $\gamma_1, \gamma_2$. As previously stated, $\gamma_1$ and $\gamma_2$ enable the trade-off between the data-fit and regularisation. Molchanov \textit{et al.}~\cite{molchanov2017variational} approximated the $\mathcal{KL}$ divergence between the prior and the posterior using $\lambda=\frac{\sigma^2}{\mu^2}$ as $\mathcal{KL}(.) \approx k_1 \sigma(k_2 + k_3 \log \lambda)-0.5\log(1+\lambda^{-1}) - k_1;\ k_1= 0.63576, k_2=1.8732, k_3 = 1.48695$. After the search or training of the final evaluation, the variances are only considered to compute which weights can be pruned and otherwise they are not considered during evaluation.

Additionally, we are inspired by~\cite{chu2019fair} which promotes the confidence in selecting connections  in a graph by explicitly minimising their entropy $\mathcal{H}$ in a similar NAS setup to minimise their uncertainty. In our case, we want to achieve high level of certainty in the operations' selection across $\boldsymbol{\alpha}^{i,j}$, which is equivalent to minimising their joint entropy across the potential operations $K$ as $\sum_{i,j}\mathcal{H}(\mathbb{E}_{q(\boldsymbol{\alpha}\mid \boldsymbol{\theta}_\alpha)}[\boldsymbol{z}^{i,j}])$. Applying a regulated coefficient $\gamma_3$ on the entropy term, the final search objective $\mathcal{L}(.)$ is formulated in \eqref{eq:search_objective_2}.
\begin{multline}
    \mathcal{A} \Leftarrow \boldsymbol{\theta}_{\boldsymbol{\alpha}}^{*} = \textrm{arg} \min_{\boldsymbol{\theta}_\alpha, \boldsymbol{\theta}_w, \boldsymbol{\Psi}} \mathcal{L}(\boldsymbol{\theta}_\alpha, \boldsymbol{\theta}_w, \boldsymbol{\Psi}) = \\ = -\mathbb{E}_{q(\boldsymbol{\alpha},\boldsymbol{w})}[\log p(\boldsymbol{y}|\boldsymbol{x}, \boldsymbol{\alpha}, \boldsymbol{w}, \boldsymbol{\Psi})] + \\ + \gamma_1 \sum_{i,j,k,c} \mathcal{KL}(q(\boldsymbol{w}_{k,c}^{i,j}\mid \boldsymbol{\theta}_w) || p(\boldsymbol{w}_{k,c}^{i,j}))
    + \\ + \gamma_2 \sum_{i,j} \mathcal{KL}(q(\boldsymbol{\alpha}^{i,j}\mid \boldsymbol{\theta}_\alpha) || p(\boldsymbol{\alpha}^{i,j})) + \\ + \gamma_3 \sum_{i,j}\mathcal{H}(\mathbb{E}_{q(\boldsymbol{\alpha}\mid \boldsymbol{\theta}_\alpha)}[\boldsymbol{z}^{i,j}]) + const.
    \label{eq:search_objective_2}
\end{multline}

\subsection{Algorithm}\label{sec:vinnas_algorithm}
Our algorithm, shown in Algorithm~\ref{alg:vinnas}, is based on SGD and relies on complete differentiation of all the operations. \textit{VINNAS} iterates between two stages: (1, lines 6-8) optimisation of  $\boldsymbol{\theta}_w$ and $\boldsymbol{\Psi}$, and (2, lines 10-14) optimisation of $\boldsymbol{\theta}_\alpha$. The usage of this two-stage optimisation aims to avoid over-adaption of parameters as suggested in~\cite{liu2018darts}.
\begin{algorithm}
\caption{\textit{VINNAS}}
\label{alg:vinnas}
\begin{algorithmic}[1]
 \State Initialise $\boldsymbol{\mu}_{w}, \boldsymbol{\mu}_{\alpha}, \log\boldsymbol{\sigma}^2_{w}, \log\boldsymbol{\sigma}^2_{\alpha}$\;
 \State Initialise scaling factors $\gamma_1, \gamma_2, \gamma_3 = 0$\;
 \State Initialise $error=\infty$
 \For{$epoch$ in search budget}
 \State \textbf{Stage (1)}
 \State Sample one $batch$ for updating $\boldsymbol{\theta}_w, \boldsymbol{\Psi}$ from $\mathcal{D}_{\boldsymbol{\theta}_w, \boldsymbol{\Psi}}$
 \State Compute loss $\mathcal{L}_{\boldsymbol{\theta}_w, \boldsymbol{\Psi}}$ based on \eqref{eq:search_objective_2} with respect to $batch$
 \State  Update $\boldsymbol{\theta}_w, \boldsymbol{\Psi}$ by gradient descent: $\boldsymbol{\theta}_w \leftarrow \boldsymbol{\theta}_w - \nabla_{\boldsymbol{\theta}_w}\mathcal{L}_{\boldsymbol{\theta}_w, \boldsymbol{\Psi}}$; $\boldsymbol{\Psi} \leftarrow \boldsymbol{\Psi} - \nabla_{\boldsymbol{\Psi}}\mathcal{L}_{\boldsymbol{\theta}_w, \boldsymbol{\Psi}}$\;
  \State \textbf{Stage (2)}
  \If{$epoch \geq weight\ epochs$}
  \State Sample one $batch$ for updating $\boldsymbol{\theta}_\alpha$ from $\mathcal{D}_{\boldsymbol{\theta}_\alpha}$\;
  \State Compute loss $\mathcal{L}_{\boldsymbol{\theta}_\alpha}$ based on \eqref{eq:search_objective_2} with respect to $batch$\;
  \State Update $\boldsymbol{\theta}_{\alpha}$ by gradient descent: $\boldsymbol{\theta}_{\alpha} \leftarrow \boldsymbol{\theta}_{\alpha} - \nabla_{\boldsymbol{\theta}_{\alpha}}\mathcal{L}_{ \boldsymbol{\theta}_{\alpha}}$\;
  \EndIf
  \State \textbf{end if}
  \State Compute error on $\mathcal{D}_{\boldsymbol{\theta}_\alpha}$
  \If{Error on $\mathcal{D}_{\boldsymbol{\theta}_\alpha} <$ $error$}
  \State Save $\boldsymbol{\theta}_\alpha$ and update $error$
  \EndIf
  \State \textbf{end if }
   \State Linearly increase $\gamma_1, \gamma_2, \gamma_3$\;
 \EndFor
 \State \textbf{end for}
 \State Choose $\mathcal{A}$ based on the positive signal to noise ratio $\frac{\boldsymbol{\mu}_\alpha}{\boldsymbol{\sigma}^2_\alpha}$ 
\end{algorithmic}
\end{algorithm}
After the initialisation of the parameters, the optimisation loops over stages (1) and (2) using two same-sized portions of the dataset. The optimisation of the stage (2) is not started from the very beginning, but only after a certain number of epochs - \textit{weight epochs}, which are used as a warm-up for training the weights of the individual operations, to avoid oscillations and settling in local minima~\cite{liu2018darts}. The variance parameters are optimised as logarithms to guarantee computational stability. We linearly increase the values of $\gamma_1,\gamma_2$ and $\gamma_3$ to force the cells to gradually choose the most relevant operations and weight patterns with respect to $\boldsymbol{\theta}_\alpha, \boldsymbol{\theta}_w$ and $\boldsymbol{\Psi}$. To avoid stranding into a local minima, we do not enforce the regularisation from the very start of the search, meaning the $\gamma$s are initialised as zero. After each iteration of (1) and (2), we compute the error on the data sampled from $\mathcal{D}_{\boldsymbol{\theta}_\alpha}$ and save the $\boldsymbol{\theta}_\alpha$ if that error was lower than that in  previous iterations. The search is repeated until the search budget, which is defined as the number of epochs that the search is allowed to perform, is not depleted. Note that the parameters for the weights $\boldsymbol{\theta}_w$ or $\boldsymbol{\Psi}$ are discarded after the search. The main outcome of the search algorithm is the parameters $\boldsymbol{\theta}_\alpha$ that are used further to perform the architecture selection that leads to $\mathcal{A}$.

Signal to noise ratio (SNR) is a commonly used measure in signal processing to distinguish between useful information and unwanted noise contained in a signal. In the context of NN architecture, the SNR can be used as an indicative of parameter importance; the higher the SNR, the more effective or important the parameter is to the model predictions for a given task. In this work we propose to look at the SNR when choosing the operations through the learnt variances $\boldsymbol{\sigma}^2_\alpha$, which can be used to compute the positive SNR as $\frac{\boldsymbol{\mu}_\alpha}{\boldsymbol{\sigma}^2_\alpha}$. We consider positive SNR, due to sign-sensitive softmax with respect to which the means were inferred. It can then be used as a metric based on which the right operations should be chosen, instead of just relying on the means $\boldsymbol{\mu}_\alpha$ as in the previous work~\cite{liu2018darts}.

\begin{table*}[t]
\centering
\caption{Comparison of found architectures from VINNAS to random search.}
\label{tab:comp_random}
\scalebox{1.}{
\setlength\tabcolsep{6pt} 
\begin{tabular}{c|c|c|c|c}
\toprule
\multirow{2}{*}{\bf Dataset} & \multirow{2}{*}{\bf Method} & {\bf Test Accuracy} (\%) &  {\bf \# Params} (M)& {\bf Search Cost} \\
& &  Positive SNR $\qquad \qquad\ $  Magnitude & Positive SNR $\qquad \qquad\ $  Magnitude & {(GPU days)}  \\
\midrule
\multirow{2}{*}{\bf MNIST} &\textit{VINNAS} & $99.47 \pm 0.07\ \boldsymbol{(99.57)}$ $\mid$ $99.47 \pm 0.07\ \boldsymbol{(99.54)}$ &  $0.01 \pm 0.002 \ \boldsymbol{(0.01)}$ $\mid$ $0.01 \pm 0.002 \ \boldsymbol{(0.01)}$   &  0.02  \\
 & Random & $97.86 \pm 1.43\ \boldsymbol{(99.5)}$&  $0.01 \pm 0.003\ \boldsymbol{(0.01)}$ &  0.0 \\
\midrule
\multirow{2}{*}{\bf FashionMNIST} & \textit{VINNAS} &   $96.02 \pm 0.10\ \boldsymbol{(96.14)}$ $\mid$ $95.97 \pm 0.12\ \boldsymbol{(96.13)}$ & $ 1.65 \pm 0.42\ \boldsymbol{(1.98)}$ $\mid$ $ 1.59 \pm 0.41\ \boldsymbol{(1.65)}$ &  0.46  \\
& Random &  $95.97 \pm 0.15\ \boldsymbol{(96.12)}$ & $ 1.74 \pm 0.70\ \boldsymbol{(1.90)}$ &  0.0  \\
\midrule
\multirow{2}{*}{\bf CIFAR-10} & \textit{VINNAS} &  $95.70 \pm 0.40\ \boldsymbol{(96.06)}$ $\mid$ $95.61 \pm 0.31\ \boldsymbol{(96.06)}$ & $ 2.17 \pm 0.58\ \boldsymbol{(1.77)}$ $\mid$ $ 1.91 \pm 0.65\ \boldsymbol{(2.00)}$  &  0.81  \\
& Random&   $94.81 \pm 0.56\ \boldsymbol{(95.81)}$ & $ 1.27 \pm 0.52\ \boldsymbol{(1.64)}$ &  0.0  \\
\bottomrule
\end{tabular}}
\end{table*}

\begin{table}
  \caption{Comparison on NAS methods for CIFAR-10.}
  \label{tab:cifar_10}
  \centering
  \scalebox{.9}{
  \begin{tabular}{c|c|c|c|c}
    \toprule
    \begin{tabular}[x]{@{}c@{}}\textbf{Search}\\\textbf{method}\end{tabular} &
    \begin{tabular}[x]{@{}c@{}}\textbf{Principal}\\\textbf{algorithm}\end{tabular}&
    \begin{tabular}[x]{@{}c@{}}\textbf{Test Accuracy}\\ (\%)\end{tabular}&
    \begin{tabular}[x]{@{}c@{}}\textbf{\# Params}\\ (M)\end{tabular}& \begin{tabular}[x]{@{}c@{}}\textbf{Search Cost}\\(GPU days)\end{tabular} \\ 
    
    \midrule
    Yamada \textit{et al.}~\cite{yamada2016deep} & hand-made  & 97.33 & 26.2 & -     \\
    Li \& Talkwalkar~\cite{li2019random} & random  & 97.15& 4.3 & 2.7     \\
    Liu \textit{et al.}~\cite{liu2018darts} & gradient  & 97.24$\pm$0.09& 3.4 & 1     \\
    Zoph \textit{et al.}~\cite{zoph2016neural} & reinf. lear. & 97.35 & 3.3 & 1800 \\
    Real \textit{et al.}~\cite{real2019regularized} & genetic alg. & 97.45$\pm$0.05 & 2.8 & 3150 \\
    Liu \textit{et al.}~\cite{cai2018path} & Bayesian opt. & 96.59$\pm$0.09 & 3.2 & 225 \\
    Zhou \textit{et al.}~\cite{zhou2019bayesnas} & gradient  & 97.39$\pm$0.04 & 3.40$\pm$0.62 & \textbf{0.2} \\
    Chu \textit{et al.}~\cite{chu2020noisy} & gradient  & 97.61 & 3.25 & - \\
    Chu \textit{et al.}~\cite{chu2019fair} & gradient  &  \textbf{97.46$\pm$0.05} & 3.32$\pm$0.46 & - \\
    Zela \textit{et al.}~\cite{zela2019understanding} & gradient & 97.05 &  - & - \\
    \midrule
    \textit{VINNAS} [Ours] & gradient & 96.06 & \textbf{1.77} & 0.81   \\
    \bottomrule
  \end{tabular}}
\end{table}
\section{Experiments}\label{sec:experiments}
To demonstrate the effectiveness of the proposed VINNAS method, we perform experiments on three different datasets, namely MNIST (M), FashionMNIST (F) and CIFAR-10 (C). 
\subsection{Experimental Settings}\label{sec:experiments_settings}
For each dataset, we search for a separate architecture involving operations commonly used in CNNs, namely: $\mathcal{O}= \{$ $3 \times3$, $5 \times 5$ and $7 \times 7$ separable convolutions, $3 \times 3$ and $5 \times 5$ dilated separable convolutions, $ 7 \times 1$ followed by $1\times 7$ convolution, $3 \times 3$ max pooling, $3 \times 3$ average pooling, skip-connection, and zero - meaning no connection$\}$ making $K=10$. Note that we clip the strength of the zero operation to avoid scaling problems with respect to other operations. All operations are followed by BN and ReLU activation except zero and skip-connection. 

Each cell accepts an input from the previous cells $c-1$ and $c-2$. Each input is processed trough ReLU-convolution-BN block to match the input shape required by that particular cell. For M, we search for an architecture comprising of a single reduction cell (R), $l=1$ with $I=2$ states.
For F, we search for an architecture comprising of 6 normal (N) and 2 reduction cells $l=2$ (NNRNNRNN) with $I=3$ states each. Both of these architectures have the same layout during evaluation, however, for F, the number of channels is increased by a factor 6.4 during evaluation. For C, during the search phase we optimise a network consisting of 8 cells $l=2$ with $I=4$ states (NNRNNRNN) that is then scaled to 20 cells during evaluation (6NR6NR6N), along with the channel sizes, which are increased by threefold. Each state always accepts 2 inputs processed through 2 operations. Each net also has a stem, which is a $3\times 3$ convolution followed by BN.
At the end of the network, we perform average pooling followed by a linear classifier with the softmax activation. Scaling of the found architectures and the followed building principles are based on previous successful work~\cite{elsken2017simple}.

The search space complexity for each net is given as $K^{(\sum_{i=0}^{I}2+i)\times l}$ which for M is $\approx 10^{5}$, for F is $\approx 10^{18}$ and for C is $\approx 10^{28}$. Weights learnt from the search phase are not kept and we retrain the resultant architectures from scratch. We train the networks with respect to a single sample with respect to $q(.)$s and LRT.
Instead of cherry-picking of the found architectures through further evaluation and then selecting the resultant architectures by hand~\cite{liu2018darts}, we report the results of the found architectures directly through \textit{VINNAS}. 

\paragraph{Search Settings} For optimising both the architecture parameters as well as the weight parameters, we use Adam~\cite{kingma2014adam} with different initial learning rates. We use cosine scheduling~\cite{loshchilov2016sgdr} for the learning rate of the weights' parameters and we keep the architecture's learning rate constant through the search. We initialise $\gamma$s and start applying and gradually linearly increasing them during the search process. We disable tracking of BN's learnable parameters for affine transformation or stats tracking. We initialise the operations strengths' $\boldsymbol{\mu}_\alpha$ through sampling $\mathcal{N}(0., 0.001)$. We utilise label smoothing~\cite{muller2019does} to avoid the architecture parameters to hard commit to a certain pattern. To speed up the search we not only search reduced architectures in terms of the number of channels and cells, but also search on 25\% - M, 50\% - F and 50\% - C of the data, while using 50\% of that portion as the dataset for learning the architecture parameters. For M we use z-normalisation.
For F and C we use random crops, flips and erasing~\cite{zhong2017random}, together with input channel normalisation. We search for 20, 50 and 100 epochs for M, F and C respectively. 

\paragraph{Evaluation Settings} During evaluation we scale up the found architectures in terms of channels and cells as described previously. We again use Adam optimiser with varying learning rates and cosine learning rate scheduling. We similarly initialise $\gamma_1$ and start to linearly increase it from a given epoch. We do so, to avoid over-regularisation and clamping of the weights to zero too soon during the optimisation. We train on full datasets for M, F and C for 100, 400 and 600 epochs respectively, and we preserve the data augmentation strategies also during retraining, we add drop-path~\cite{larsson2016fractalnet} and auxiliary tower~\cite{szegedy2015going} regularisation to C and F. For both the search and evaluation we initialise the weights' means with Xavier uniform initialisation~\cite{glorot2010understanding}. We initialise all the log-variances to $-10$. 
\begin{figure}
    \centering
    \includegraphics[width=0.9\linewidth]{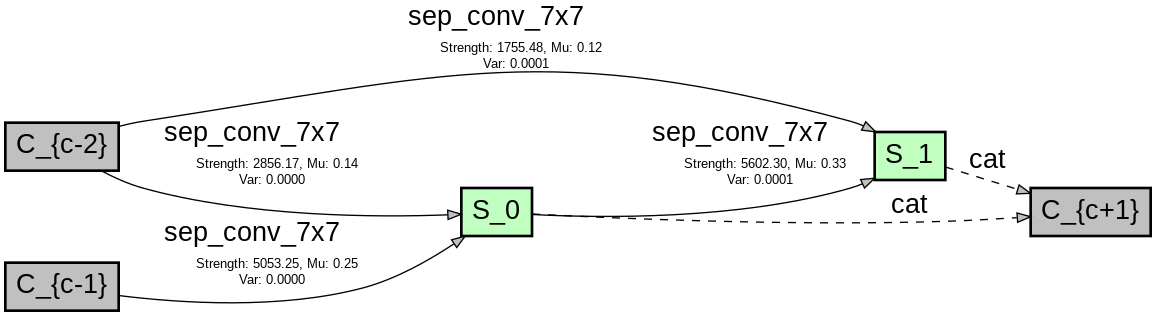}
    \caption{MNIST reduction cell with its positive SNR.}
    \label{fig:mnist}
\end{figure}
\begin{figure}
    \centering
    \includegraphics[width=0.45\linewidth]{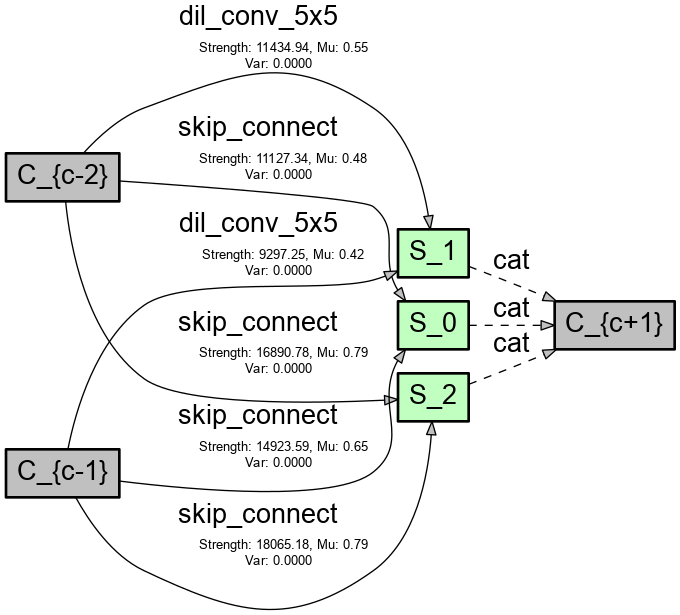}
    \includegraphics[width=.9\linewidth]{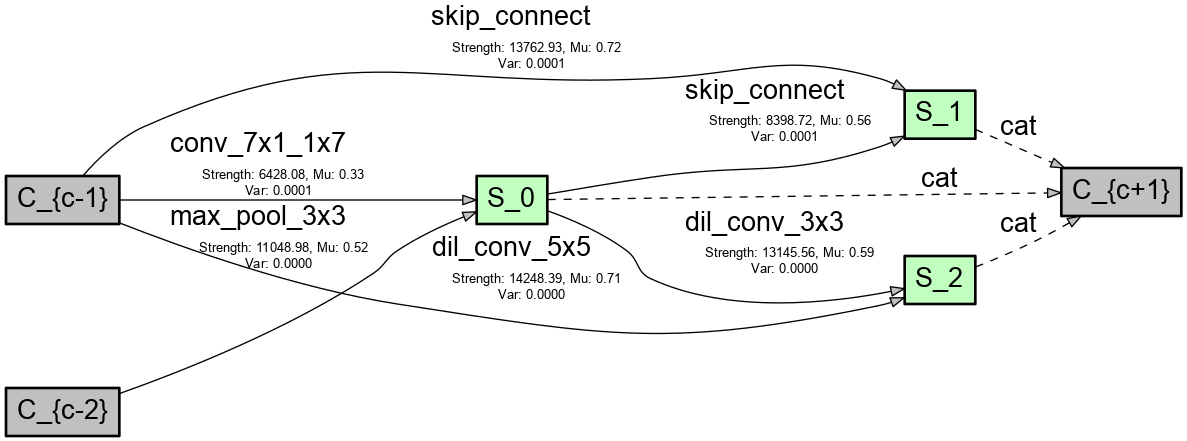}
    \caption{FashionMNIST normal and reduction cells with their positive SNR.}
    \label{fig:fashion}
\end{figure}
\begin{figure}
    \centering
    \includegraphics[width=1\linewidth]{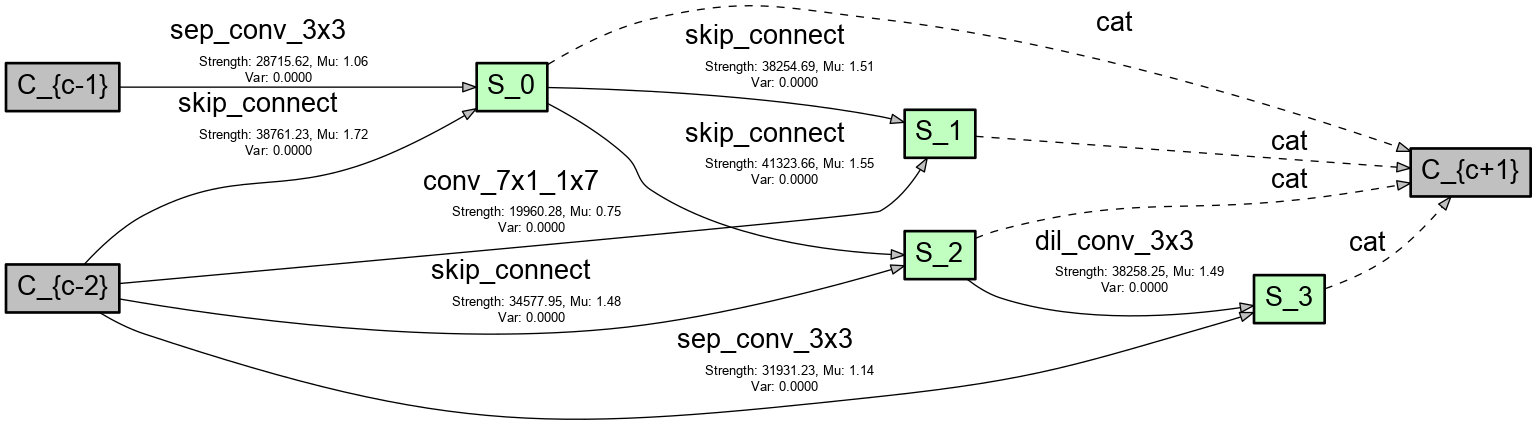}
    \includegraphics[width=1\linewidth]{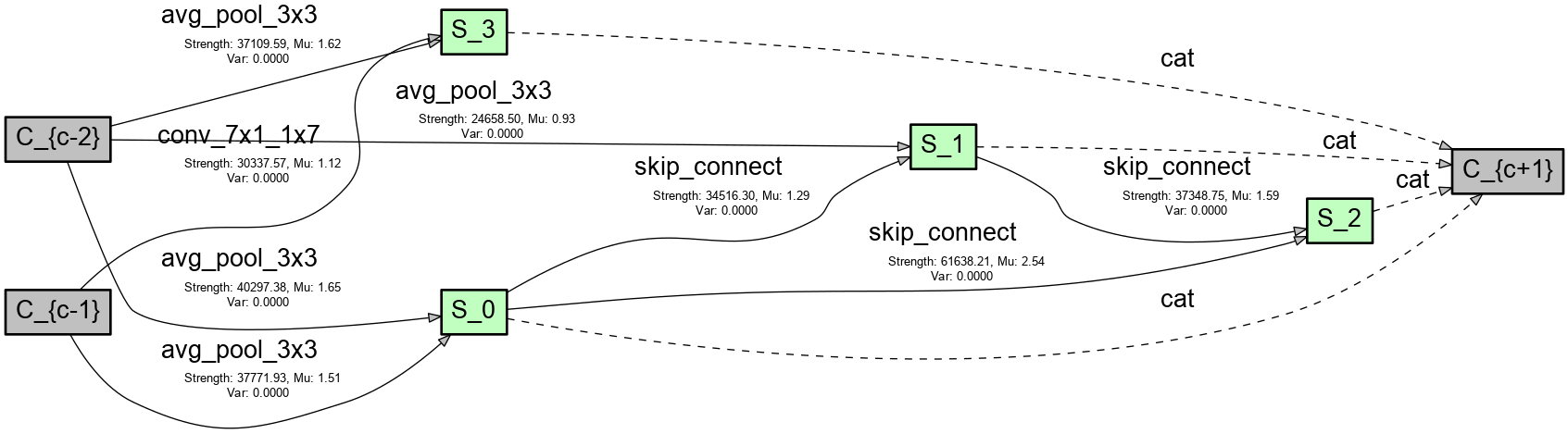}
    \caption{CIFAR-10 normal and reduction cells with their positive SNR.}
    \label{fig:cifar}
\end{figure}

\subsection{Evaluation}\label{sec:experiments_evaluation}
The evaluation is condensed in Tables~\ref{tab:comp_random} and~\ref{tab:cifar_10}. The numbers in bold represent the score for the best performing model for the given selection method: positive SNR/magnitude and the dataset. The found best performing architectures are shown in Figures~\ref{fig:mnist},~\ref{fig:fashion} and~\ref{fig:cifar}. Specifically for the case of CIFAR-10, that is popular in the NAS community, in Table~\ref{tab:cifar_10} it is shown that \textit{VINNAS} found an architecture that is comparable to the SOTA, however, with $2 \times$ fewer non-zero parameters.

We first perform random search on our search spaces for M, F and C. Note that the search spaces are vast and we deem it impossible to evaluate all architectures in the search space, given our available resources, and thus we sample 10 separate architectures from each search space and we train them with the same hyperparameter settings as the found architectures to avoid any bias. The number of parameters for \textit{VINNAS} is reported as the amount after pruning with respect to $\log \lambda \geq 3$.

When comparing the found architectures for the different datasets in Table~\ref{tab:comp_random}, we noticed that in case of M, there are certain connections onto which an operation could potentially be completely omitted with the positive SNR being relatively small. We attribute this to the fact that this dataset is easy to generalise to, which can be also seen by the overall performance of the random search for these datasets. However, on CIFAR-10, it can be seen that the inferred importance of all the operations and the structure is very high. The results also demonstrated that using the learnt uncertainty in the operation selection, in addition to the magnitude, marginally benefits the operation selection. Compared with DARTS~\cite{liu2018darts} which only uses $3 \times 3$ separable convolutions and max pooling everywhere, it can be observed that the found architectures are rich in the variety of operations that they employ and the search does not collapse into a mode where all the operations are the same. For future reference regarding deeper models such as for F and C, we observe that the found cells of the best performing architectures do contain skip-connections to enable efficient propagation of gradients and better generalisation.

The main limiting factor of this work is the GPU search cost which is higher, in comparison to the other NAS methods, due to using LRT, which requires two forward passes during both search and evaluation. Most importantly, all the found architectures demonstrate good generalisation performance in terms of the measured test accuracy.

\section{Conclusion}\label{sec:conclusion}

In summary, our work proposes a combined approach of probabilistic modelling and neural architecture search. Specifically, we give the operations' strengths a probabilistic interpretation by viewing them as learnable random variables. Automatic relevance determination-like prior is imposed on these variables, along with their corresponding operation weights, which incentivises automatic detection of pertinent operations and zeroing-out the others. Additionally, we promote certainty in the operations selection, through a custom loss function which allows us to determine the most relevant operations in the architecture. We demonstrated the effectiveness of \textit{VINNAS} on three different datasets and search spaces. 

In the future work, we aim to explore a hierarchical Bayesian model for the architecture parameters, which could lead to architectures composed of more diverse cell types, instead of just two. Additionally, all of the evaluated NNs shared the same evaluation hyperparameters and in the future we want to investigate an approach which can automatically determine suitable hyperparameters for the found architecture.

\bibliographystyle{ieeetr}
\bibliography{bib.bib}

\appendix

\section{Derivation of the Search Objective}\label{sec:appx_search_objective}

The search objective presented in~\eqref{eq:search_objective_2} is derived in a following way, starting from the $\mathcal{KL}$ divergence between our approximation and the true posterior $\mathcal{KL}(q(\boldsymbol{\alpha}, \boldsymbol{w}\mid\boldsymbol{\theta}_\alpha, \boldsymbol{\theta}_w) \mid \mid p(\boldsymbol{\alpha}, \boldsymbol{w}, \boldsymbol{\Psi} \mid \boldsymbol{x}, \boldsymbol{y}))$.
\begin{multline}
    \mathcal{KL}(q(\boldsymbol{\alpha}, \boldsymbol{w}\mid\boldsymbol{\theta}_\alpha, \boldsymbol{\theta}_w) \mid \mid p(\boldsymbol{\alpha}, \boldsymbol{w}, \boldsymbol{\Psi} \mid \boldsymbol{x}, \boldsymbol{y})) \\
    = \int\int q(\boldsymbol{\alpha}, \boldsymbol{w}\mid\boldsymbol{\theta}_\alpha, \boldsymbol{\theta}_w) \log\frac{q(\boldsymbol{\alpha}, \boldsymbol{w}\mid\boldsymbol{\theta}_\alpha, \boldsymbol{\theta}_w)}{p(\boldsymbol{\alpha}, \boldsymbol{w}, \boldsymbol{\Psi} \mid \boldsymbol{y}, \boldsymbol{x})} d\boldsymbol{\alpha}d\boldsymbol{w}
\end{multline}

Then we use the Bayes rule to decompose the posterior $p(\boldsymbol{\alpha}, \boldsymbol{w}, \boldsymbol{\Psi} \mid \boldsymbol{x}, \boldsymbol{y}) = \frac{p(\boldsymbol{y}\mid \boldsymbol{x}, \boldsymbol{\alpha}, \boldsymbol{w}, \boldsymbol{\Psi})p(\boldsymbol{\alpha})p(\boldsymbol{w})p(\boldsymbol{\Psi})}{p(\boldsymbol{y}\mid\boldsymbol{x})}$. We can separate the priors $p(\boldsymbol{\alpha}, \boldsymbol{w},\boldsymbol{\Psi})$ into individual terms $p(\boldsymbol{\alpha})p(\boldsymbol{w})p(\boldsymbol{\Psi})$ as they are independent from one another in our model.
\begin{multline}\hspace{-0.5cm}
    = \int\int q(\boldsymbol{\alpha}, \boldsymbol{w}\mid\boldsymbol{\theta}_\alpha, \boldsymbol{\theta}_w) \log\frac{q(\boldsymbol{\alpha}, \boldsymbol{w}\mid\boldsymbol{\theta}_\alpha, \boldsymbol{\theta}_w)p(\boldsymbol{y}\mid\boldsymbol{x})}{p(\boldsymbol{y}\mid \boldsymbol{x}, \boldsymbol{\alpha}, \boldsymbol{w}, \boldsymbol{\Psi})p(\boldsymbol{\alpha})p(\boldsymbol{w})p(\boldsymbol{\Psi})} \\ d\boldsymbol{\alpha}d\boldsymbol{w}
\end{multline}

Since the marginal term $p(\boldsymbol{y}\mid \boldsymbol{x})$ is independent of the parameters it can be moved outside of the integral where it is a constant along with the term $p(\boldsymbol{\Psi})$, which we assume is a uniform prior.
\begin{multline}
    = \int\int q(\boldsymbol{\alpha}, \boldsymbol{w}\mid\boldsymbol{\theta}_\alpha, \boldsymbol{\theta}_w) \log\frac{q(\boldsymbol{\alpha}, \boldsymbol{w}\mid\boldsymbol{\theta}_\alpha, \boldsymbol{\theta}_w)}{p(\boldsymbol{y}\mid \boldsymbol{x}, \boldsymbol{\alpha}, \boldsymbol{w}, \boldsymbol{\Psi})p(\boldsymbol{\alpha})p(\boldsymbol{w})}\\
    d\boldsymbol{\alpha}d\boldsymbol{w} +const.
\end{multline}

Next, we separate the terms with respect to the logarithm, into two terms: one involving the log-likelihood with respect to the data $p(\boldsymbol{y}\mid \boldsymbol{x}, \boldsymbol{\alpha}, \boldsymbol{w}, \boldsymbol{\Psi})$ and the other which consists of the priors $p(\boldsymbol{\alpha})p(\boldsymbol{w})p(\boldsymbol{\Psi})$ and the approximate posterior $q(\boldsymbol{\alpha}, \boldsymbol{w}\mid\boldsymbol{\theta}_\alpha, \boldsymbol{\theta}_w)$. Then, since not only the priors $p(\boldsymbol{\alpha}),p(\boldsymbol{w})$, but also the approximations $q(\boldsymbol{\alpha}\mid\boldsymbol{\theta}_\alpha),q(\boldsymbol{w}\mid\boldsymbol{\theta}_w)$ are independent, we can split the integral between $q(\boldsymbol{\alpha}\mid\boldsymbol{\theta}_\alpha)$ and $q(\boldsymbol{w}\mid\boldsymbol{\theta}_w)$ which again result in two $\mathcal{KL}$ divergence terms, in addition to the log-likelihood.
\begin{multline}
     = -\mathbb{E}_{q(\boldsymbol{\alpha},\boldsymbol{w}\mid  \boldsymbol{\theta}_\alpha, \boldsymbol{\theta}_w)}[\log p(\boldsymbol{y}\mid \boldsymbol{x}, \boldsymbol{\alpha}, \boldsymbol{w}, \boldsymbol{\Psi})] +\\ \qquad \int\int q( \boldsymbol{\alpha}\mid \boldsymbol{\theta}_\alpha) q(\boldsymbol{w}\mid\boldsymbol{\theta}_w) \log\frac{q( \boldsymbol{\alpha}\mid \boldsymbol{\theta}_\alpha) q(\boldsymbol{w}\mid\boldsymbol{\theta}_w)}{p(\boldsymbol{\alpha})p(\boldsymbol{w})}\\ d\boldsymbol{\alpha}d\boldsymbol{w} +const. 
\end{multline}
\begin{multline}
   = -\mathbb{E}_{q(\boldsymbol{\alpha},\boldsymbol{w}\mid  \boldsymbol{\theta}_\alpha, \boldsymbol{\theta}_w)}[\log p(\boldsymbol{y}\mid \boldsymbol{x}, \boldsymbol{\alpha}, \boldsymbol{w}, \boldsymbol{\Psi})] + \\ \qquad \mathcal{KL}( q(\boldsymbol{w}\mid\boldsymbol{\theta}_w)\mid\mid p(\boldsymbol{w})) + \mathcal{KL}( q(\boldsymbol{\alpha}\mid\boldsymbol{\theta}_\alpha)\mid\mid p(\boldsymbol{\alpha}))\\ +const.
\end{multline}
$\mathbb{E}_{q(\boldsymbol{\alpha},\boldsymbol{w}\mid  \boldsymbol{\theta}_\alpha, \boldsymbol{\theta}_w)}[\log p(\boldsymbol{y}\mid \boldsymbol{x}, \boldsymbol{\alpha}, \boldsymbol{w}, \boldsymbol{\Psi})]$ represents the log-likelihood with respect to the samples from the approximates $q(\boldsymbol{\alpha},\boldsymbol{w}\mid  \boldsymbol{\theta}_\alpha, \boldsymbol{\theta}_w)$ and the data, which in our case is the standard cross-entropy term. Note, that the expectation is thus approximated through Monte Carlo sampling with respect to these variables and also the data $\boldsymbol{x}, \boldsymbol{y}$. The weights $\boldsymbol{w}$ as well as the architecture weights $\boldsymbol{\alpha}$ are independent for each operation $o^{i,j}_{k,c}(.)$ and therefore the $\mathcal{KL}$ divergence can be computed independently for each term resulting in sums indexed by $i,j,k,c$.
\begin{multline}
    = -\mathbb{E}_{q(\boldsymbol{\alpha},\boldsymbol{w} \mid \boldsymbol{\theta}_\alpha, \boldsymbol{\theta}_w})[\log p(\boldsymbol{y}|\boldsymbol{x}, \boldsymbol{\alpha}, \boldsymbol{w}, \boldsymbol{\Psi})] \\ + \sum_{i,j,k,c} \mathcal{KL}(q(\boldsymbol{w}_{k,c}^{i,j}\mid \boldsymbol{\theta}_w) || p(\boldsymbol{w}_{k,c}^{i,j}))
     \\ + \sum_{i,j} \mathcal{KL}(q(\boldsymbol{\alpha}^{i,j}\mid \boldsymbol{\theta}_\alpha) || p(\boldsymbol{\alpha}^{i,j})) + const.
\end{multline}

Furthermore, we introduced arbitrary constants $\gamma_1$ and $\gamma_2$ to balance the effect of the regulariser terms $\mathcal{KL}(.)$. Note, that we compute the $\mathcal{KL}$ divergence with respect to the approximation provided by Molchanov \textit{et al.}~\cite{molchanov2017variational}.
\begin{multline}
    = -\mathbb{E}_{q(\boldsymbol{\alpha},\boldsymbol{w} \mid \boldsymbol{\theta}_\alpha, \boldsymbol{\theta}_w)}[\log p(\boldsymbol{y}|\boldsymbol{x}, \boldsymbol{\alpha}, \boldsymbol{w}, \boldsymbol{\Psi})] \\ + \gamma_1 \sum_{i,j,k,c} \mathcal{KL}(q(\boldsymbol{w}_{k,c}^{i,j}\mid \boldsymbol{\theta}_w) || p(\boldsymbol{w}_{k,c}^{i,j}))
    \\ + \gamma_2 \sum_{i,j} \mathcal{KL}(q(\boldsymbol{\alpha}^{i,j}\mid \boldsymbol{\theta}_\alpha) || p(\boldsymbol{\alpha}^{i,j})) + const.
\end{multline}

Lastly, we add the entropy term $\mathcal{H}$ to increase the certainty of the operations' selection. In our case, we want to achieve certainty in the operations' selection across $\boldsymbol{\alpha}^{i,j}$, which is equivalent to minimising their joint entropy across the potential operations $K$ as $\sum_{i,j}\mathcal{H}(\mathbb{E}_{q(\boldsymbol{\alpha}\mid \boldsymbol{\theta}_\alpha)}[\boldsymbol{z}^{i,j}])$. The $\boldsymbol{z}^{i,j}$ are computed with respect to the samples from $q(\boldsymbol{\alpha} \mid \boldsymbol{\theta}_\alpha)$ in \eqref{eq:darts}. Applying a regulating coefficient $\gamma_3$ on the entropy term gives the final search objective. 
\begin{multline}
    = -\mathbb{E}_{q(\boldsymbol{\alpha},\boldsymbol{w} \mid \boldsymbol{\theta}_\alpha, \boldsymbol{\theta}_w)}[\log p(\boldsymbol{y}|\boldsymbol{x}, \boldsymbol{\alpha}, \boldsymbol{w}, \boldsymbol{\Psi})] + \\ + \gamma_1 \sum_{i,j,k,c} \mathcal{KL}(q(\boldsymbol{w}_{k,c}^{i,j}\mid \boldsymbol{\theta}_w) || p(\boldsymbol{w}_{k,c}^{i,j}))
    + \\ + \gamma_2 \sum_{i,j} \mathcal{KL}(q(\boldsymbol{\alpha}^{i,j}\mid \boldsymbol{\theta}_\alpha) || p(\boldsymbol{\alpha}^{i,j})) + \\ + \gamma_3 \sum_{i,j}\mathcal{H}(\mathbb{E}_{q(\boldsymbol{\alpha}\mid \boldsymbol{\theta}_\alpha)}[\boldsymbol{z}^{i,j}]) + const.
\end{multline}

The same logic, but with fewer terms, can be applied to derive the original ELBO in~\eqref{eq:elbo}.

Below in Tables~\ref{tab:mnist} and \ref{tab:fashion} we present the comparison of \textit{VINNAS} with respect to other related hand-made and NAS-found architectures for MNIST and FashionMNIST datasets.
\begin{table}[H]
  \caption{Comparison of VINNAS for MNIST.}
  \label{tab:mnist}
  \centering
  \scalebox{.9}{
  \begin{tabular}{c|c|c|c|c}
    \toprule
    \begin{tabular}[x]{@{}c@{}}\textbf{Search}\\\textbf{method}\end{tabular} &
    \begin{tabular}[x]{@{}c@{}}\textbf{Principal}\\\textbf{algorithm}\end{tabular}&
    \begin{tabular}[x]{@{}c@{}}\textbf{Test Accuracy}\\ (\%)\end{tabular}&
    \begin{tabular}[x]{@{}c@{}}\textbf{\# Params}\\ (M)\end{tabular}& \begin{tabular}[x]{@{}c@{}}\textbf{Search Cost}\\(GPU days)\end{tabular} \\ 
    
    \midrule
    LeCun \textit{et al.}~\cite{lecun1998gradient} & hand-made  & 99.45 & 0.37 & -     \\
    Jin \textit{et al.}~\cite{jin2019auto} & Bayes. opt.  & 99.45 & -& 0.5     \\
    Fedorov \textit{et al.}~\cite{fedorov2019sparse} & Bayes. opt.  & 99.17 & 0.001& 1     \\
    Byla \textit{et al.}~\cite{byla2019deepswarm} & swarm. opt.  & \textbf{99.61} & -& 0.33    \\
    Gaier \textit{et al.}~\cite{gaier2019weight} & genetic alg.  & 91.9 & $\approx$ \textbf{0}& -   \\
    
    \midrule
    \textit{VINNAS} [Ours] & gradient & 99.57 & 0.01 & \textbf{0.02}  \\
    \bottomrule
  \end{tabular}}
\end{table}
\begin{table}[H]
  \caption{Comparison of VINNAS for FashionMNIST.}
  \label{tab:fashion}
  \centering
  \scalebox{.9}{
  \begin{tabular}{c|c|c|c|c}
    \toprule
    \begin{tabular}[x]{@{}c@{}}\textbf{Search}\\\textbf{method}\end{tabular} &
    \begin{tabular}[x]{@{}c@{}}\textbf{Principal}\\\textbf{algorithm}\end{tabular}&
    \begin{tabular}[x]{@{}c@{}}\textbf{Test Accuracy}\\ (\%)\end{tabular}&
    \begin{tabular}[x]{@{}c@{}}\textbf{\# Params}\\ (M)\end{tabular}& \begin{tabular}[x]{@{}c@{}}\textbf{Search Cost}\\(GPU days)\end{tabular} \\ 
    
    \midrule
   Zhong \textit{et al.}~\cite{zhong2017random} & hand-made  & $96.2 \pm 0.05$ & 11 & -     \\
    Nøkland \& Eidnes~\cite{nokland2019training} & hand-made  & 95.47& 7.3& -     \\
    Jin \textit{et al.}~\cite{jin2019auto} & Bayes. opt.  & 92.58 & -& 0.5     \\
    Kyriakides \textit{et al.}~\cite{10.1007/978-3-030-49186-4_10} & genetic alg.  & 94.46 & 3.1& -     \\
    Byla \textit{et al.}~\cite{byla2019deepswarm} & swarm. opt.  & 93.56 & -& 0.33    \\
    Xue \textit{et al.}~\cite{xue2019transferable} & clustering  & 93.9 & -& \textbf{0.013 }    \\
    Noy \textit{et al.}~\cite{noy2020asap} & gradient  & 96.27 & -& 0.2    \\
    
    Nayman \textit{et al.}~\cite{nayman2019xnas} & gradient  & 96.36 & 3.7& 0.3     \\
    Tanveer \textit{et al.}~\cite{tanveer2020fine} & gradient  & \textbf{96.91} & 3.2& -    \\

    \midrule
    \textit{VINNAS} [Ours] & gradient & 96.14 & \textbf{1.98} & 0.46   \\
    \bottomrule
  \end{tabular}}
\end{table}

As can be seen in the Tables~\ref{tab:mnist} and~\ref{tab:fashion}, our method is comparable to the state-of-the-art results in terms of accuracy as well as the number of non-zero parameters. \textit{VINNAS} can find an architecture with a comparable performance to other works for classifying MNIST digits as well as FashionMNIST images. These results prove the versatility of our method, which can be used for finding CNN architectures for simple and more challenging tasks alike.

\end{document}